\pdfoutput=1

\documentclass[11pt]{article}

\usepackage{graphicx}
\usepackage{wrapfig}
\usepackage{float}
\usepackage{hyphenat}
\usepackage{acl}
\usepackage{times}
\usepackage{latexsym}

\usepackage[T1]{fontenc}
\usepackage[utf8]{inputenc}

\usepackage{microtype}

\title{Disambiguation of morpho-syntactic features of African American English -- the case of habitual \textit{be}}

\author{Harrison Santiago\\ 
Department of Computer and \\Information Science and Engineering,\\University of Florida \\ \texttt{harrison.santiag@ufl.edu} \And
  Joshua L. Martin \\
 Department of Linguistics,\\ University of Florida \\  \texttt{joshua.martin@ufl.edu}\AND
 Sarah Moeller \\
 Department of Linguistics,\\ University of Florida \\  \texttt{smoeller@ufl.edu}\\\And
 Kevin Tang \\
 Department of English and \\American Studies,\\ Heinrich-Heine-University, D\"{u}sseldorf \\  \texttt{kevin.tang@hhu.de}
 \\}

\begin{document}
\maketitle
\begin{abstract}
Recent research has highlighted that natural language processing (NLP) systems exhibit a bias against
African American speakers. The bias errors are often caused by poor representation of linguistic features unique to African American English (AAE), due to the relatively low probability of occurrence of many such features in training data. We present a workflow to overcome such bias in the case of habitual ``be''. Habitual ``be'' is isomorphic, and therefore ambiguous, with other forms of ``be'' found in both AAE and other varieties of English. This creates a clear challenge for bias in NLP technologies. To overcome the scarcity, we employ a combination of rule-based filters and data augmentation that generate a corpus balanced between habitual and non-habitual instances. With this balanced corpus, we train unbiased machine learning classifiers, as demonstrated on a corpus of AAE transcribed texts, achieving .65 F$_1$ score disambiguating habitual ``be''.
\end{abstract}

\section{Introduction}

%BRIEFLY: SITUATION OF AAE AND NLP - MOTIVATION / OUR SPECIFIC PROBLEM - OUR MODEL - PREVIEW OF RESULTS - HINT AT BROADER IMPACT/IMPLICATIONS

Linguistic discrimination has adversely affected the lives of marginalized populations for centuries, including racially marginalized groups in the United States. In spite of extensive research on linguistic discrimination \cite{baugh2008linguistic}, many NLP systems inherit the linguistic biases that exist between humans. For example, preliminary studies into the performance of automatic speech recognition (ASR) systems uncovered a performance bias against African American speakers \citep{tatman2017effects,dorn2019dialect}. This problem was confirmed most recently by \citet{koenecke2020racial} who found that the average word error rate (WER) for white American speakers was significantly lower as compared to the average WER for African American speakers among five prominent ASR systems from such companies as Google, Amazon, and Apple.  

This performance gap is rooted in two related issues. First, the linguistic differences between African American English (AAE) and General American English (GAE) include distinctive features in their  morphosyntactic structures. Second, incorrect inferences in NLP systems are often caused by the scarcity of certain linguistic features when training, and the many unique features in AAE have a relatively low probability of occurrence. 

This paper describes work that overcomes the data scarcity issue for a specific feature unique to AAE: the habitual ``be''. As the name suggests, this morphologically invariant form of “be” communicates habitual action. Disambiguating habitual ``be'' from non-habitual ``be'' is difficult for two prominent reasons. First, the form is isomorphic with the other uses of ``be'', such as the infinite use in ``I want to be...''. Second, habitual ``be'' is relatively rare even in corpora of AAE. Our work addresses both these issues. It uses a rule-based method that capitalizes on morphosyntactic differences to eliminate a portion of non-habitual ``be'' instances and it uses a method of data augmentation that increases the ratio of habitual ``be'' instances. The resulting balanced data can then be used to train classifiers to tag ``be'' instances as habitual or non-habitual.\footnote{https://github.com/HarrisonSantiago/Habitual\_be\_classifier}
%While not currently available, all related code will be published at \currentlyurl{https://github.com/HarrisonSantiago/Habitual-be-classifier}}
%\textbf{CHECK IF THE URL NEEDS TO BE ANOMYNISED AS WELL. IT REVEALS WHO YOU ARE.}

%THE PROBLEM WE WANT TO SOLVE: HABITUAL BE IS SO RARE, HOW DO WE OVERCOME THE DATA SCARCITY PROBLEM? IN 3 SENTENCES OR LESS, HERE IS HOW OUR PAPER SOLVES THAT

\begin{figure*}
\label{flowchart}
\includegraphics[width=\textwidth]{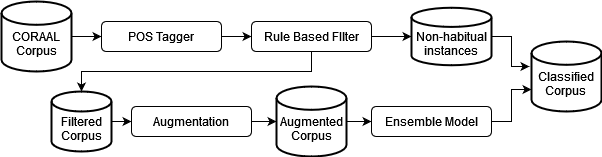}
\caption{The disambiguation pipeline: the input corpus goes through a Part-of-Speech tagger, after which non-habitual instances are separated by a rule-based filter. Any indeterminate ``be'' instances are balanced by augmentation and tagged by classification models.}
\label{flowchart}
\end{figure*}

\section{Related work}
%OVERVIEW OF RELATED PUBLISHED RESEARCH WITH HINTS HOW IT RELATES SPECIFICALLY TO OURS. e.g., WHAT WAS THE NLP TOOL/MODEL, HIGHLIGHTS OF WORK, RESULTS, WHAT WAS DONE WELL, WHAT MISSING or WHAT QUESTIONS WERE RAISED AND NOT ANSWERED
Distinguishing habitual ``be'' and non-habitual ``be'' usage is a word sense disambiguation (WSD) problem because it involves
%has a long successful history in computational linguistics \citep[][Ch. 18]{Jurafsky2016}. WSD 
identifying the meaning of words in context \citep{navigli2009word}. Most successful WSD algorithms make use of contextual embeddings 
\citep{melamud2016context2vec,peters2018deep}, but some feature extraction 
algorithms, such as the IMS algorithm by \citet{zhong-ng-2010-makes}, have a comparable level of performance although  comparatively much simpler. The IMS algorithm uses a support-vector-machine (SVM) with simple contextual features, such as word form or part-of-speech (POS) tags, and weighted average of embeddings. %which are extracted from surrounding words or n-grams. 
Similarly, our disambiguation pipeline makes use of the POS tags of the surrounding words. This helps avoid the limited amount of annotated AAE data 
%(see Sections \ref{hab.be.exs} and \ref{sec:data}) 
which could lead to sparse word vectors and unreliable embeddings. 

%and not the surrounding words or their embeddings as features 
%; furthermore, available pretrained embeddings (e.g., \citet{bojanowski2017enriching}) are inappropriate as they were not typically trained on AAE data. 

%[JUSTIFY the uSE OF RULE_based models because we are dealing iwhta syntactic phenon.]

%Our evaluation of the effects of augmentation serves a proof-of-concept that the augmented data have the potential to increase the performance of other AAE-focused NLP systems. In the future, we will evaluate our pipeline against state-of-the-art neural models such as the Transformer model.

%To address the data sparsity issue with AAE, our pipeline employs 
Data augmentation techniques that generate synthetic, or artificial, language in the training data often improve NLP applications when the training corpus is small or when a certain feature occurs rarely \citep{chen2021empirical}. Our approach follows previously successful examples of data augmentation methods that combine a language model \citep{fadaee-etal-2017-data} with a thesaurus \citep{Zhang_thesaurus} or word embeddings \citep{wu_bertaugment}. These methods identify substitutes for words in the data and insert them into synthetic strings that include the target feature. 

% Feature-Based WSD
% Feature-based algorithms for WSD are extremely simple and function almost as
% well as contextual language model algorithms. The best performing IMS algorithm
% (Zhong and Ng, 2010), augmented by embeddings (Iacobacci et al. 2016, Raganato
% et al. 2017b), uses an SVM classifier to choose the sense for each input word with
% the following simple features of the surrounding word

% Melamud et al. (2016) and Peters et al. (2018).

%  Our classification An SVM classifier to choose the sense for each input word with
% the following simple features of the surrounding words
% \citet{zhong-ng-2010-makes}
% Feature-based algorithms for WSD
% part-of-speech tags (for a window of 3 words on each side, stopping at sen-
% tence boundaries)
% collocation features of words or n-grams of lengths 1, 2, 3 at a particularcollocation
% location in a window of 3 words on each side (i.e., exactly one word to the
% right, or the two words starting 3 words to the left, and so on).
% weighted average of embeddings (of all words in a window of 10 words on
% each side, weighted exponentially by distance)

%\textbf{KEVIN TO ADD CITATIONS}
% We need a place to describe habitual be and give examples. I added a subsection here as an example. 
\section{Habitual ``be''}
\label{hab.be.exs}

The ``be'' verb has various functions. This includes several types of non-habitual use, as shown in Appendix \ref{sec:appendix:nonhabbe}. The use of habitual ``be'' is a prominent, distinct, and well-researched morphosyntactic feature in AAE. Habitual ``be'' is a morphologically invariant form of the verb that encodes the habitual aspect, as shown below \citep{green2002african}. 

\begin{enumerate}
    \item I \textbf{be} in my office by 7:30. (habitual: AAE) 
    \item I \textbf{am usually} in my office by 7:30. (habitual: GAE)
\end{enumerate}

Syntactic contexts serve as important cues for disambiguating ``be'' as habitual or non-habitual. \citet{Martin_Tang_RacialBias_HabBe_Interspeech2020_2020} show that ASR systems not only fail to recognize habitual ``be'' more often than non-habitual ``be'' but, when habitual ``be'' is present in an utterance, the surrounding words are also incorrectly recognized, particularly preceding words. These findings reveal a strong dependency between habitual ``be'' and its syntactic context. Failure to reflect this dependency in a language model could lead to a less accurate and biased system.

%was a significant predictor of the word recognition errors, and the effect of preceding context and word error was higher than in the following context. It was proposed that this is likely due to language models which only take into account the context of preceding words.
%[NOTE: COME BACK TO THIS WITH TWO THOUGHTS: ]

%HERE YOU SAY HOW WE CAN MEDIATE THIS ISSUE FOR ASR SYSTEMS. NEED TO IMPROVE THE LANGUAGE MODEL BY a) INCORPORAING MORE HABITUAL BE INSTANCES IN THE TRAINING DATA AND B) DISAMBIGUATING HAB-BE AND NON-HAB-BE IN THE TRAINING DATA FOR A MORE PRECISE LANGUAGE MODEL.

Even in an AAE corpus, habitual ``be'' is relatively rare.
%(see Section \ref{sec:data})
%compared to non-habitual ``be'',
 %SAY HOW THIS DATA DISTRIBUTION POSES A CHALLENGE FOR ANY SPEECH TECHNOLOGY SYSTEM DESIGNED FOR AAE (OR AT LEAST NOT BIASED AGAINST AAE). 
This imbalanced distribution poses a challenge for designing a non-biased NLP system because most classifiers tend to be biased towards the majority class. 

The ambiguity and scarcity of habitual ``be'' presents two obvious approaches to a solution: (i) incorporate more habitual ``be'' instances in the data, (ii) manually disambiguate habitual and non-habitual ``be'' before training. 
Each approach poses a challenge. For (i), simply collecting more data is extremely impractical, as the habitual ``be'' is naturally rare. For (ii), hand-coding is unsuitable for the scale of the data needed. 
%We need a method to automatically disambiguate morphosyntactic features of AAE over billions of words. 

Our study addresses these challenges with a rule-based filter based on syntactic cues and with a data augmentation technique. Together the filter and data augmentation increase the ratio of habitual ``be'', providing a more balanced training set for the model and allowing for a more fine-grained language model. 

\section{Methodology}

The first novel task towards training classifiers to disambiguate habitual ``be'' is to address the ambiguity of the invariant form by eliminating as many non-habitual ``be'' instances as possible. The second task is to increase the proportional occurrence of habitual ``be'' in the training data. 

We undertake these two tasks and incorporate them into a pipeline, shown in Figure \ref{flowchart}. First, the entire corpus is run through a pre-trained \texttt{NLTK} tokenizer and POS tagger trained using the Penn Treebank Project. To eliminate as many non-habitual ``be'' instances as possible, a rule-based filter identifies determinate instances of non-habitual ``be''. With these removed, we increase the proportional occurrence of habitual ``be'' by augmenting the proportion of habitual ``be''. Finally, we combine the filtered habitual ``be'' instances back into a now balanced dataset and use that dataset to train an ensemble model for classification. As discussed in section 5.1, the habituality of each instance is known and allows accurately creating rules and training the classifiers.     

\subsection{Preprocessing}

The data is formatted using WordSmith Tools \citep{WordSmithTools} so that each instance of ``be'' is centered in a 102-character string, the length being determined by the software default. To simplify the task, no breaks between speakers or texts were included, meaning these text segments combine speech from multiple speakers and texts if necessary, with no indication as to where this occurs. If multiple instances of ``be'' fall within 102 characters, each instance is treated as separate instance that becomes the center of another string slightly offset from the overlapping example. Also, all punctuation, marks made by transcribers (e.g.,  ``/??/''), corpus-specific codes (e.g., ``/RD-NAME-3/'') and other non-speech text are removed as part of the preprocessing.

\subsection{Rule-based filter of non-habitual ``be''}

In AAE, there are certain syntactic patterns that strongly correlate to occurences of the habitual ``be'' \citep{green2002african, fasold_tense_1972}.
Most patterns are based on the part-of-speech immediately surrounding ``be''. Two example patterns are a pronoun immediately preceding ``be''  (e.g., ``...\textit{they} \textbf{be} like, what you finna do?'') and a verb ending in -ing immediately following ``be'' (e.g., ``But LeBron \textbf{be} \textit{passing} though'').
% \begin{itemize}
% \item Pronoun immediately proceeding ``be''  (e.g., ``Let's see, or they \textbf{be} like, what you finna do?'')
% \item Verb ending in -ing immediately following ``be'' (e.g., ``But LeBron \textbf{be} passing though'')
% \end{itemize} 

Following from this, we invert some patterns and create filters that capture a large number of non-habitual instances. For example, if the word that precedes ``be'' is not a pronoun and the word after it is not a verb ending in -ing, then we can say that instance is non-habitual.

The vast majority of non-habitual ``be'' instances are caught by these syntactic rules. In addition, we created some ad-hoc rules that showed success at eliminating remaining non-habitual ``be'', although they generally capture a smaller number. 
A full list of our rules we can be found in Appendix \ref{sec:appendix:rulesnonhabbe}. 

The goal of the rule-based filter is not to identify instances of habitual ``be''. Rather, it is used to remove non-habitual ``be'' instances for which more advanced disambiguation techniques are not needed. This is a step towards creating a more balanced corpus. It serves to narrow the scope of our classifier to those instances which much more difficult to be automatically disambiguated. 

\subsection{Augmenting habitual ``be''}

To counter the relative rarity of habitual ``be'', the dataset needs to be balanced, but without excluding the remaining non-habitual instances after the rule-based filter is applied. Instead, the amount of habitual ``be'' can be increased.  To accomplish this, we use data augmentation to create new, synthetic examples of habitual ``be''. 

We found that the Python library \texttt{nlpaug} \citep{ma2019nlpaug} provides easy synthetic text generation. Focusing on text augmentation, we used the Word2Vec \citep{Mikolov2013EfficientEO}\footnote{https://github.com/dav/word2vec} and WordNet \citep{wordnet} implementations for substituting and inserting words in surrounding examples of habitual ``be'' instances from our corpus. The Word2Vec implementation both substitutes and inserts new words at random by finding similar words using the cosine distance from pre-trained embeddings. The WordNet augmentation leverages a database of semantic relations to substitute synonyms at random. These methods can occasionally lead to ungrammatical outputs, as seen in Appendix \ref{sec:appendix:augmentedhab}. We did not remove such occurrences, as the inclusion of all generated perturbations in our data set strengthened the robustness of our model. Combined, these methods inserted or replaced words with a new part of speech in over 90\% of the augmentations.

% Would randomly replace with synonyms for whichever is smaller: 10 words, 30% of the input string
% Randomly decides where to insert words, again either 10 words or 30% of original string size. 

% \begin{figure}[h]
% \includegraphics[width=7.5cm]{percentage1}
% \caption{Affects of data Augmentation on our training corpus. The top row shows the affect on the raw number of instances, while the bottom shows the affect on the overall percentage of habitual instances in the corpus.}
% \label{augment}
% \end{figure}

\subsection{Classifiers}

After filtering trivial instances of non-habitual ``be'' and balancing the remaining data by augmenting instances of habitual ``be'', we train a logistic regression classifier, a multi-layer perceptron (MLP), and a linear Support Vector Machine (SVM) to disambiguate instances of ``be''. All are implemented with the \texttt{scikit-learn} library. All models set the max-iteration to 10,000 steps to allow for convergence on a regular basis. The MLP was changed to use a limited-memory BFGS algorithm solver, and set to have two hidden layers, the first with five nodes and the second with two. These hyperparameters were set after a non-exhaustive search of looking for the optimal settings. All other default parameters were kept unchanged. We compared these against a majority-rules ensemble model that uses the logistic regression, MLP, and SVM voting algorithms. The votes are equally weighted between all three.

The input to all of the classifiers consists of vectors which contain the number of times each POS occurs within a window around each instance of ``be''. We treated the size of this window as a hyperparameter, and found that defining our window to start at the 9th word in the string and end at the 5th-from-last word produced optimal results.

%PROVIDE LINK to REPOSITORY HERE OR IN INTRO

\section{Experiment}

Unbiased NLP systems should successfully disambiguate instances of habitual ``be''. We implemented our system on a corpus of AAE speakers after training it our filtered and balanced corpus. %This allowed us to explore the viability of our classification pipeline.
%, and particularly to evaluate the effects of augmentation on our success.

\subsection{Data}\label{sec:data}

% From josh:  What interviews, how many hours, how much data
%SOURCE OF DATA - TYPE OF TEXTS - STATS OF TRAINING/DEV/TEST, ETC.
%KT: these were copied from Martin and Tang
%the Corpus of Regional African American Language (CORAAL) \cite{kendall2018corpus}. CORAAL contains over 100 sociolinguistic interviews with African American speakers, totaling to over 105 hours of audio and including a rich variety of interviewees that vary widely by age, socio-economic background, gender identity, and urban/ruralness. 

%3,635 instances of the word ``be'' were collected from transcripts of the interviews. Each instance of ``be'' was hand-tagged as habitual/non-habitual and instances of ``be'' from non-Black interviewers were filtered out, resulting in 376 instances of habitual ``be'' and 2,974 instances of non-habitual ``be''.  

The data comes from the Corpus of Regional African American Language (CORAAL) \citep{kendall2018corpus} which contains transcriptions of over 150 sociolinguistic interviews with African American speakers, totaling more than 127 hours of audio and including a rich variety of interviewees by age, socio-economic background, gender identity, and urban/rural origin. 

%3,635 instances of the word ``be'' were collected from transcripts of the interviews. Each instance of ``be'' was hand-tagged as habitual/non-habitual and instances of ``be'' from non-Black interviewers were filtered out, resulting in 376 instances of habitual ``be'' and 2,974 instances of non-habitual ``be'' 

\begin{table}
\centering
\begin{tabular}{c|c|c|c}
%\hline
%\multicolumn{3}{|c|}{Effects of Augmentation}                                                                                        \\ \hline
\multicolumn{1}{c|}{}                                                                         & \multicolumn{1}{c|}{Orig.} & Filter & Augment  \\ \hline
\multicolumn{1}{c|}{\begin{tabular}[c]{@{}c@{}}Non-hab ``be'' total \end{tabular}}  & 4,656 & \multicolumn{1}{c|}{994}    & 944 \\ \hline
\multicolumn{1}{c|}{\begin{tabular}[c]{@{}c@{}}Hab ``be'' total \end{tabular}}  & 477 & \multicolumn{1}{c|}{416}    & 963 \\ \hline
\multicolumn{1}{c|}{\begin{tabular}[c]{@{}c@{}}Hab `be'' \%  \end{tabular}} & 9\% & \multicolumn{1}{c|}{30\%}   & 50\%  \\ %\hline
\end{tabular}
\caption{The distribution of habitual ``be'' in the training corpus:  original, rule-based filtered, and augmented. The top two rows show the change in the raw number of ``be'' instances; the bottom shows the proportion of habitual ``be'' to non-habitual ``be''.}
\label{tab:augment}
\end{table}

From this corpus, 5,133 instances of ``be'' were manually annotated as habitual/non-habitual. This resulted in 477 instances of habitual ``be'' and, 4,656 instances of non-habitual ``be'', which is to say that non-habitual instances were approximately ten times more frequent. The rule-based filter and augmentation were applied to this data with the resulting statistics shown in Table \ref{tab:augment}. The rule-based filter incorrectly eliminated 61 instances of habitual ``be'', reducing the total from 477 to 416. This means the filter has an error rate of about 13\% that might be improved with additional ad-hoc rules.

%CORAAL features time-aligned orthographic transcriptions of recorded speech in numerous locations across the United States. In particular 
%we looked at …. (INFO FROM JOSH).  

%Some preprocessing was done to our input corpus in order to avoid features which were not a result of speech. All punctuation was removed, as it would be unnecessary for any of our classifying components. Similarly, transcription marks such as ``/??/'', substrings such as “/RD-NAME-3/” and other non-speech text were removed. Finally, some common contractions were expanded to aid our POS tagger.

% Some preprocessing was done to our input corpus in order to avoid features which were not a result of speech. All punctuation was removed, as it would be unnecessary for any of our classifying components. We wished to remove transcription marks such as “/??/”. Similarly, substrings such as “/RD-NAME-3/” and other non-speech text were removed. Finally, some common contractions were expanded to aid our POS tagger.  

When analyzing our classifiers, we used a 70/30 training/test split, with the test set having a ratio of non-habitual to habitual occurrences similar to that of the original corpus. Importantly, the dataset was split before any augmentation occurred to help our results be more transferable to the original corpus. To get a better understanding of the consistency in results that the augmentation methods would lead to, we re-performed our augmentation procedure for each trial. In total, 10 trials were performed. %We believe that this found a balance between including relevant speech while excluding other speakers. 

\subsection{Results}

% \begin{figure}[h]
% \includegraphics[width=7.5cm]{table1(1)}
% \caption{Table of F-scores for different classification algorithms (Logistic regression, Support Vector Machine, Multilayer Perceptron, and Ensemble method). The mean over 10 trials are reported for each, with the standard deviation accompanying in parentheses.}
% \label{results}
% \end{figure}

%TABLE/CHART OF RESULTS. DISCUSS
Based on our results on the CORAAL corpus, classifying habitual ``be'' is a feasible task even with a limited supply of natural AAE speech for training. Each algorithm and the ensemble model were tested after being trained on the filtered and the augmented data and on the original corpus. Table \ref{results} shows F$_1$-scores displays the comparison, showing means and standard deviations over 10 trials. The best results were achieved by the ensemble classifier after both filtering and augmenting. Over 10 trials the ensemble model classified instances of habitual ``be'' with an average score of 0.65. 

All four classifiers' performance rose dramatically when using our filtering and augmentation methods. In addition, the variability in classifier performance decreased after filtering and augmentation, as evident by the lower standard deviations. The lower variability indicates that balancing a data set allowed the classifiers to find a more definitive decision boundary.

\begin{table}
\begin{tabular}{ccc}
%\hline
%\multicolumn{3}{|c|}{F$_1$-scores}                                                                    \\ \hline
\multicolumn{1}{c|}{}                    & \multicolumn{1}{c|}{\begin{tabular}[c]{@{}c@{}}Augmented \end{tabular}} & \begin{tabular}[c]{@{}c@{}}Not Augmented\end{tabular} \\ \hline
\multicolumn{1}{c|}{\begin{tabular}[c]{@{}c@{}}Logistic\\ regression\end{tabular} } & \multicolumn{1}{c|}{0.648 (0.048)}                                               & 0.416 (0.039)                                                  \\ \hline
\multicolumn{1}{c|}{SVM}                 & \multicolumn{1}{c|}{0.628 (0.114)}                                               & 0.542 (0.206)                                                  \\ \hline
\multicolumn{1}{c|}{MLP}                 & \multicolumn{1}{c|}{0.627 (0.038)}                                               & 0.498 (0.058)                                                  \\ \hline
\multicolumn{1}{c|}{Ensemble}            & \multicolumn{1}{c|}{\textbf{0.652} (0.049)}                                               & 0.439 (0.084)                                                  \\ %\hline
\end{tabular}
\caption{F$_1$-scores for different classification algorithms (Logistic regression, Support Vector Machine (SVM), Multilayer Perceptron (MLP), and Ensemble of all three). The mean over 10 trials are reported, with the standard deviation in parentheses.}
\label{results}
\end{table}

\section{Conclusion}

%RECAP BRIEFLY AND INTRODUCE (or recap) FUTURE STEPS OR QUESTIONS THAT STILL NEED TO BE ADDRESSED. ALSO, BROADER IMPLICATIONS AND IMPACTS
Our goal was to develop a pipeline which aids the creation of models unbiased against African American English. We proposed and tested a combination of hand-crafted rules, data augmentation, and machine learning to disambiguate instances of habitual ``be'' which is a distinct, if relatively infrequent, morphosyntactic feature in AAE. The results show this combination to be a promising pipeline, with each step contributing to success at increasing classification scores and reducing bias. 

The hand-crafted rules we used took into consideration morphosyntactic patterns that are unique to AAE and correlate with habitual ``be'' usage. This allowed us to filter out most non-habitual ``be'' instances. We then found that Word2Vec and WordNet augmentation methods were able to adequately imitate AAE structure and balance the proportion of habitual ``be'' instances. Together the filtering and the augmentation resulted in more balanced data with which to train the classifiers. %Together these methods proved an effective way to improve a system's ability to disambiguate habitual from non-habitual ``be''. 

In the future, with an increased amount of natural speech and more advanced classification algorithms, it is possible that the classification performance could be even higher. However, due to limited data, we treated the entire CORAAL corpus without regard to several interesting factors that should be considered. For example, we did not regard the geographic location or origin of the speaker. Further analysis of our model's performance with respect to regional sub-varieties of AAE would be an interesting avenue to explore. This exploration might refine the hand-crafted rules. Also, our pipeline makes use of the POS tags of the surrounding words, similar to \citep{zhong-ng-2010-makes}, but it does not include the surrounding words themselves or their embeddings as features because the limited data would have led to sparse word vectors and unreliable embeddings.  
%Our evaluation of the effects of augmentation serves a proof-of-concept that the augmented data have the potential to increase the performance of other AAE-focused NLP systems. In the future, we will evaluate our pipeline against state-of-the-art neural models such as the Transformer model.

We feel it should be easy to adapt our pipeline to other unique AAE features such as the completive ``done'' \citep{green2002african}.
Although we expect feature-based models to tend to perform better at low-resource settings than deep learning, we plan to compare our results against state-of-the-art neural models such as the Transformer \citep{vaswani2017attention}.
 %If this works, it would be a step towards assembling the necessary corpus for language models unbiased towards AAE.

The increase in scores we were able to achieve with these simple methods serves as a proof-of-concept that systems based on similar syntactic filtering and data augmentation approaches have the potential to improve the performance of other AAE-focused NLP systems and provide enough data for more advanced feature representations. %such as word embeddings.  

%%% REFERENCES %%%%%
% Entries for the entire Anthology, followed by custom entries
\bibliography{references}
\bibliographystyle{acl_natbib}

\appendix

\section{Appendix: Types of non-habitual ``be''}
\label{sec:appendix:nonhabbe}

\begin{itemize}
\item auxiliary ``be'' in progressive constructions (e.g., ``I will \textbf{be} going there tomorrow.'')
\item auxiliary ``be'' in passive constructions (e.g., ``She should \textbf{be} given an award.'')
\item copula or auxiliary ``be'' preceded by verbal complements (e.g., ``He wanted to \textbf{be} a lawyer.'')
\item copula or auxiliary ``be'' preceded by a modal (e.g., ``They might \textbf{be} in the house.'')
\item imperative ``be'' (e.g., ``\textbf{Be} quiet!'')
\end{itemize}

\section{Rules to filter non-habitual ``be''}
\label{sec:appendix:rulesnonhabbe}

\begin{itemize}
\item If the word immediately preceding ``be'' is a modal, adjective, or ``to''.
\item If the word immediately following ``be'' is a verbal noun, while the word immediately preceding is not a personal pronoun nor a noun.
\item If the word immediately following ``be'' is an adjective, while the word immediately preceding ``be'' is not a personal pronoun nor a noun.
\item If the word immediately following ``be'' is a preposition or subordinating conjunction, while the word immediately preceding ``be'' is a singular present verb.
\item If the word immediately preceding ``be'' is a noun, and the word immediately preceding that noun is an adjective
\item If the word immediately preceding ``be'' is an adverb, and  the word immediately following ``be'' is either a personal pronoun or determiner.
\item If the word immediately preceding ``be'' is an adverb, and either the word immediately preceding the adverb is a verb, or modal
\end{itemize}

\section{Examples of augmenting occurrences of the habitual ``be''}
\label{sec:appendix:augmentedhab}

\begin{itemize}
\item "they were like you should totally come here we be having so much fun So I tell my mom about it and" becomes "they were like you should totally come hither we be have got so much fun So I tell my mom astir it and"

\item "mixed up all kinds a way everybody just just be there having a good time That s Mm hm that s" becomes "mixed up all dizzying array a way everybody yeah just be happen having a heckuva time That s hm that s"
\end{itemize}

\end{document}